\pdfoutput=1
\documentclass[11pt]{article}

\usepackage[final]{ACL2023}

\usepackage{times}
\usepackage{latexsym}

\usepackage[T1]{fontenc}
\usepackage[utf8]{inputenc}

\usepackage{microtype}
\usepackage{inconsolata}

\usepackage[T1]{fontenc}

\usepackage{times}
\usepackage{latexsym}
\usepackage{danudefs}
\usepackage{algorithmic}
\usepackage{algorithm}
\usepackage{color}
\usepackage{booktabs}
\usepackage{xspace}
\usepackage{url}
\usepackage{color, colortbl}
\definecolor{Gray}{gray}{0.9}
\usepackage{subfigure}
\usepackage[export]{adjustbox}
\usepackage{makecell}
\usepackage[english]{babel}
\usepackage{hyperref}
\addto\captionsenglish{%
}
\addto\extrasenglish{%
}

\newcommand{\red}[1]{\textcolor{red}{#1}}
\newcommand{\blue}[1]{\textcolor{blue}{#1}}

\title{Learning Dynamic Contextualised Word Embeddings via \\Template-based Temporal Adaptation}

\author{Xiaohang Tang$^\dagger$ \And
    Yi Zhou$^\ast$  \\
  University of Liverpool$^\dagger$, Cardiff University$^\ast$, Amazon$^\ddagger$\\
  {\tt \{sgxtang4, danushka\}@liverpool.ac.uk} \\
  {\tt zhouy131@cardiff.ac.uk}\And
  Danushka Bollegala$^{\dagger,\ddagger}$}

\date{}

\begin{document}
\maketitle

\begin{abstract}
Dynamic contextualised word embeddings (DCWEs) represent the temporal semantic variations of words.
We propose a method for learning DCWEs by time-adapting a pretrained Masked Language Model (MLM) using time-sensitive templates.
Given two snapshots $C_1$ and $C_2$ of a corpus taken respectively at two distinct timestamps $T_1$ and $T_2$, we first propose an unsupervised method to select (a) \emph{pivot} terms related to both $C_1$ and $C_2$, and (b) \emph{anchor} terms that are associated with a specific pivot term in each individual snapshot.
We then generate prompts by filling manually compiled templates using the extracted pivot and anchor terms.
Moreover, we propose an automatic method to learn time-sensitive templates from $C_1$ and $C_2$, without requiring any human supervision.
Next, we use the generated prompts to adapt a pretrained MLM to $T_2$ by fine-tuning using those prompts.
Multiple experiments show that our proposed method reduces the perplexity of test sentences in $C_2$, outperforming the current state-of-the-art.
\end{abstract}

\section{Introduction}
\label{sec:intro}

\iffalse
Words, the elementary meaning-bearing constituents of a language, often change their meaning over time~\cite{Baybee:2015,Rama:2015}.
The meaning of a word can depend on extralinguistic contexts such as time~\cite{Koch:2016} and social space~\cite{Robinson:2010}.
For example, the word \emph{gay} has gradually changed its meaning from \emph{happy} to \emph{homosexual} over the last five decades~\cite{Robinson:2012,Campbell:2004}.

Word meaning representations in NLP have been largely limited to representing the meaning of a word in a given snapshot of a corpus, sampled at a specific timestamp.
For example, static word embedding methods~\cite{Milkov:2013,Glove,bojanowski-etal-2017-enriching} represent the meaning of a word using a single vector in all of its mentions, ignoring sense or temporal semantic variations.
Although sense embedding methods~\cite{LMMS,ARES} learn multiple vectors corresponding to the distinct senses of a word, they do not reflect the temporal aspects related to semantic variations.
\fi

Contextualised word embeddings produced by MLMs~\cite{devlin-etal-2019-bert,RoBERTa,ALBERT,XLNet} represent the meaning of a word with respect to the context in which it appears in and have reported substantial performance gains in various NLP tasks.
The usage of words change over time and the same word might be associated with different words to mean different concepts over time~\cite{Koch:2016,Baybee:2015,Rama:2015}.
For example, the word \emph{gay} has gradually changed its meaning from \emph{happy} to \emph{homosexual} over the last five decades~\cite{Robinson:2012,Campbell:2004}.
However, MLMs are often trained using a static snapshot of a corpus taken at a specific timestamp, and are \emph{not updated} afterwards.
Because of this reason, existing pretrained MLMs do \emph{not} capture the temporal semantic variations of words. 
For example, \newcite{Loureiro:2022} showed that neither the original version of BERT~\cite{devlin-etal-2019-bert} nor RoBERTa~\cite{RoBERTa} are up-to-date with the information related to the current coronavirus pandemic.

To address the above-mentioned limitations, we propose a Dynamic Contextualised Word Embedding (\textbf{DCWE}) method that \emph{adapts} a given pretrained MLM from one  timestamp $T_1$ to another $T_2$ using two snapshots of a corpus $C_1$ and $C_2$, sampled respectively at times $T_1$ and $T_2$.
We represent a word $x$ by an embedding that depends both on the \textbf{context} $c$ of $x$, as well as on \textbf{time} $T$.
Our word embeddings are \emph{dynamic} because they depend on the time, and \emph{contextualised} because they also depend on the context.

We model the problem of adapting a given pretrained MLM to a specific timestamp $T_2$ as an instance of prompt-based fine-tuning~\cite{2107.13586}, which has been successfully used in prior work to adapt MLMs to various tasks such as relation representation~\cite{ushio-etal-2021-distilling,fichtel2021prompt}, domain adaptation~\cite{PADA}, natural language inference~\cite{utama-etal-2021-avoiding} and question answering~\cite{softprompts}.
Compared to fine-tuning MLMs on manually labelled training instances, which might not be readily available or costly to manually annotate in sufficient quantities for a particular task to fine-tune a large-scale MLM, prompt-based methods require only a small number of prompts~\cite{le-scao-rush-2021-many}. 
Luckily, in our case of temporal adaptation of MLMs~\cite{Agarwal:2022}, such prompts could be generated from a handful of manually designed templates (\autoref{sec:manual}) or automatically extracted from unlabelled data (\autoref{sec:auto}).
This aspect of our proposed method is particularly attractive compared to prior work (see \autoref{sec:related}) on DWEs~\cite{Rudolph:2018tg,hofmann-etal-2021-dynamic,HistBERT,Loureiro:2022} that require retraining of MLMs from scratch to incorporate the time-sensitive constraints into the embedding spaces.

We first extract \emph{pivot} words, $w$, that are common to both $C_1$ and $C_2$.
Second, we extract \emph{anchor} words $u$ and $v$ that are strongly associated with $w$ in respectively $C_1$ and $C_2$.
We then propose several methods to score tuples $(w,u,v)$ such that the semantic variation of $w$ from $T_1$ to $T_2$ is captured by its association with respectively $u$ and $v$.
Finally, we generate a large number of textual prompts using the top-scoring tuples $(w,u,v)$ according to each method to fill the slots in manually written templates such as
``\textit{$\langle w \rangle$ is associated with $\langle u \rangle$ in $\langle T_1 \rangle$, whereas it is associated with $\langle v \rangle$ in $\langle T_2 \rangle$.}''
Here, the slots corresponding to $T_1$ and $T_2$ are filled by specific years when respectively $C_1$ and $C_2$ were sampled.
We differentiate \emph{templates} from \emph{prompts} throughout the paper where the latter is formed by filling one or more slots in the former. 
We further propose a method to automatically generate templates from sentences selected from $C_1$ and $C_2$ using a text-to-text transformation model~\cite{JMLR:v21:20-074}, thereby obviating the need to manually create templates.
Finally, the given MLM is adapted to $T_2$ by fine-tuning it on the generated prompts.

Experimental results conducted on Reddit, Yelp, ArXiv and Ciao datasets show that the proposed prompt-based time-adapting of MLMs consistently outperforms previously proposed DCWEs~\cite{hofmann-etal-2021-dynamic} and temporal adaptation methods~\cite{TempoBERT} reporting better (lower) perplexity scores on unseen test sentences in $C_2$.
The source code for our proposed method is publicly available.\footnote{\url{https://github.com/LivNLP/TimeAdapted-DCWE}}

\section{Related Work}
\label{sec:related}

Methods that use part-of-speech~\cite{mihalcea-nastase-2012-word}, entropy~\cite{10.1007/s11280-014-0316-y}, latent semantic analysis~\cite{Sagi_2011} and temporal semantic indexing~\cite{inproceedings} have been proposed for detecting changes in word meanings.
In SemEval-2020 Task 1~\cite{schlechtweg-etal-2020-semeval} two subtasks were proposed for detecting lexical semantic change: a binary classification task (for a given set of target words, decide which words had their meaning altered, and which ones not) and a ranking task (rank a set of target words according to their degree of lexical semantic change between the two corpora).
\newcite{giulianelli-etal-2020-analysing} showed that contextualised embeddings obtained from an MLM can be used to measure the change of word meaning.
\newcite{rosin-radinsky-2022-temporal} proposed a temporal attention mechanism by extending the self-attention mechanism in transformers, where time stamps of the documents are considered when computing the attention scores.
\newcite{Aida:ACL:2023} proposed a method to predict semantic change of words by comparing the distributions of contextualised embeddings of the word between two given corpora, sampled at different points in time.
Our goal in this paper extends beyond the detection of a subset of words with a change in lexical semantics, and to adapt MLMs over time.

DWEs~\cite{Rudolph:2018tg,hofmann-etal-2021-dynamic,HistBERT,Loureiro:2022} incorporate extralinguistic information such as time, demographic or social aspects of words with linguistic information.
\newcite{welch-etal-2020-compositional} learnt demographic word embeddings, covering attributes such as age, gender, location and religion. 
\newcite{Zeng2017-rh} learnt \emph{socialised} word embeddings considering both a social media user's personal characteristics of language use and that user's social relationships.
However, \newcite{hofmann-etal-2021-dynamic} showed that temporal factors have a stronger impact than socio-cultural factors when determining the semantic variations of words.
Consequently, in this paper we focus on the temporal adaptation of DCWEs.

Diachronic Language Models that capture the meanings of words at a particular point in time have been trained using historical corpora~\cite{HistBERT,Loureiro:2022}.
These prior work learn independent word embedding models from different corpora.
This is problematic because information related to a word is not shared across different models resulting in inefficient learning, especially when word occurrences within a single snapshot of a corpus are too sparse to learn accurate embeddings.

\newcite{Rudolph:2018tg} proposed a dynamic Bernoulli embedding method based on exponential family embeddings, where each word is represented by a one-hot vector with dimensionality set to the vocabulary size.
This model is extended to the temporal case by considering different time-slices where only the word embedding vector is time-specific and the context vectors are shared across the corpus and over time-slices.
Because the joint distribution over time and context is intractable, they maximise the pseudo log-likelihood of the conditional distribution for learning the parameters of their DWE model.
\newcite{PADA} proposed a domain adaptation method based on automatically learnt prompts.
Given a test example, they generate a unique prompt and conditioned on it, then predict labels for test examples.
Although their method uses prompts to adapt a model, they do \emph{not} consider temporal adaptation of MLMs, which is our focus.
Moreover, we do not require any labelled examples in our proposal.

\newcite{Dynamic-Language-Models-for-Continuously-Evolving-Content} proposed a model updating method using vocabulary composition and data sampling to adapt language models to continuously evolving web content. 
However, their work is specific to one dataset and two classification tasks, and focuses on incremental training.
\newcite{jang2022temporalwiki} introduced a benchmark for ever-evolving language models, utilising the difference between consecutive snapshots of datasets, to track language models' ability to retain existing knowledge while incorporating new knowledge at each time point. 
%Nevertheless, they did not propose methods to handle the issue and to be compared with our approach.
\newcite{jin-etal-2022-lifelong} studied the lifelong language model pretraining problem, where the goal is to continually update pretrained language models using emerging data. 
%As they focused on pretraining process, this work cannot be directly compared with our approach.
\newcite{Time-Aware} introduced a diagnostic dataset to investigate language models for factual knowledge that changes over time and proposed an approach to jointly model texts with their timestamps. 
They also demonstrated that models trained with temporal context can be adapted to new data without retraining from scratch. 
%However, their work focused on their own introduced dataset and Question Answering tasks, which means that it is difficult to compare their approach's performance with ours.   
\newcite{TempoBERT} proposed TempoBERT, where they insert a special time-related token to each sentence and fine-tune BERT using a customised time masking. 
TempoBERT reports superior results in SemEval 2020 Task 1 semantic variation detection benchmark.
As shown later in \autoref{sec:results}, our proposed method outperforms TempoBERT.

\newcite{hofmann-etal-2021-dynamic} proposed DCWEs, which are computed in two stages.
First, words are mapped to dynamic type-level representations considering temporal and social information.
The type-level representation of a word is formed by combining a non-dynamic embedding of a word and a dynamic offset that is specific to the social and temporal aspects of the word.
Second, these dynamic embeddings are converted to context-dependent token-level representations. 
To the best of our knowledge, this is the only word embedding method that produces both dynamic as well as contextualised representations, thus mostly relates to us.
As shown in \autoref{sec:exp}, our proposed method outperforms their DCWEs on four datasets.

\section{Prompt-based Time Adaptation}
\label{sec:method}

Given two snapshots $C_1$ and $C_2$ of a corpus taken respectively at timestamps $T_1$ and $T_2 (>T_1)$, we consider the problem of adapting a pretrained MLM $M$ from $T_1$ to $T_2$.
We refer to a word $w$ that occurs in both $C_1$ and $C_2$ but has its meaning altered between the two snapshots as a \textbf{pivot}.
We propose three methods for selecting tuples $(w,u,v)$, where $u$ is closely associated with the meaning of $w$ in $C_1$, whereas $v$ is closely associated with the meaning of $w$ in $C_2$.
We name $u$ and $v$ collectively as the \textbf{anchors} of $w$, representing its meaning at $T_1$ and $T_2$.
If the meaning of $w$ has changed from $T_1$ to $T_2$, it will be associated with different sets of anchors, otherwise by similar sets of anchors.
We then fine-tune $M$ on prompts generated by substituting $(w,u,v)$ in templates created either manually (\autoref{sec:manual}) or automatically (\autoref{sec:auto}).

\subsection{Prompts from Manual Templates}
\label{sec:manual}

In order to capture temporal semantic variations of words, we create the template \textit{$\langle w \rangle$ is associated with $\langle u \rangle$ in $\langle T_1 \rangle$, whereas it is associated with $\langle v \rangle$ in $\langle T_2 \rangle$.}\footnote{We experimented with multiple manual templates as shown in the Supplementary but did not observe any clear improvements over this template.}
We generate multiple prompts from this template by substituting tuples $(w, u, v)$ extracted using three methods as described in \autoref{sec:tuple-selection}.
For example, given a tuple (\emph{mask}, \emph{hide}, \emph{vaccine}) and $T_1 = 2010$ and $T_2 = 2020$, the previous template produces the prompt: \textit{mask is associated with hide in 2010, whereas it is associated with vaccine in 2020}.
These prompts are used in \autoref{sec:FT} to fine-tune an MLM to adapt it to $T_2$ for obtaining DCWEs. 

\subsection{Tuple Selection Methods}
\label{sec:tuple-selection}

Given a template with slots corresponding to $w$, $u$ and $v$, we propose three different criteria for selecting tuples to fill those slots.

\subsubsection{Frequency-based Tuple Selection}
\label{sec:freq}

Prior work on domain adaptation has shown that words highly co-occurring in both source and target domains are ideal candidates for adapting a model trained on the source domain to the target domain.
Following prior work on cross-domain representation learning, we call such words as pivots~\cite{Bollegala:ACL:2015}. 
Specifically, we measure the suitability of a word $w$, $\mathrm{score}(w)$ as a pivot by \eqref{eq:freq}. 
\begin{align}
\label{eq:freq}
\mathrm{score}(w) = \min(f(w, C_1), f(w, C_2))
\end{align}
Here, $f(w, C_1)$ and $f(w, C_2)$ denote the frequency of $w$ respectively in $C_1$ and $C_2$, measured by the number of sentences in which $w$ occurs in each corpus.
We sort words in the descending order of the scores given by \eqref{eq:freq} and select the top $k$-ranked words as pivots.

Next, for each pivot $w$, we select its anchors $x$ by the Pointwise Mutual Information, $\mathrm{PMI}(w, x; C)$, computed from the snapshot $C$ as in \eqref{eq:pmi}.
\begin{align}
    \label{eq:pmi}
    \mathrm{PMI}(w, x; C) = \log \left( \frac{ p(w, x)}{p(w) p(x)} \right)
\end{align}
Here, $p(x)$ is the marginal probability of $x$ in $C$, estimated as $f(x, C)/N_C$, where $N_C$ is the total number of sentences in $C$.
Moreover, $p(w,x)$ is the joint probability between $w$ and $x$, estimated as ${\rm cooc}(w,x)/N_C$, where $\mathrm{cooc}$ is the total number of co-occurrences between $w$ and $x$ in $C$, considering sentences as the contextual window for the co-occurrences.

We select the set of words $\cU(w)$ with high $\mathrm{PMI}(w,u;C_1)$ values as the \emph{anchors} of $w$ in $C_1$.
Likewise, the set of words $\cV(w)$ with the top- $\mathrm{PMI}(w,v;C_2)$ are selected as the anchors of $w$ in $C_2$.
By construction, anchors are the words that are strongly associated with a pivot in each snapshot of the corpus, thus can be regarded as representing the meaning carried by the pivot in a snapshot according to the distributional hypothesis~\cite{Firth:1957}.
Finally, for each $w$, we obtain a set of tuples, $\cS_{\rm freq} = \{(w,u,v) | u \in \cU(w), v \in \cV(w)\}$, by considering all pairwise combinations of anchors with a pivot for the purpose of filling the templates to generate prompts.

\subsubsection{Diversity-based Tuple Selection}
\label{sec:NN}

Recall that our objective is to select anchors $u$ and $v$, respectively in $C_1$ and $C_2$ such that the change of meaning of a pivot $w$ is captured by the tuple $(w,u,v)$.
Frequency-based tuple selection method described in \autoref{sec:freq} finds $u$ and $v$, which are strongly associated with $w$ in the two snapshots of the corpus.
However, if $\cU(w)$ and $\cV(w)$ are highly similar, it could mean that the meaning of $w$ might not have changed from $T_1$ to $T_2$.
%because according to the distributional hypothesis~\cite{Firth:1957} $w$ is associated with almost the same set of neighbours in the two snapshots of the corpus.
To address this issue, we define \emph{diversity} of $w$ as the dissimilarity between its sets of anchors as in \eqref{eq:diversity}.
\begin{align}
    \label{eq:diversity}
    \mathrm{diversity}(w) = 1 - \frac{|\cU(w) \cap \cV(w)|}{|\cU(w) \cup \cV(w)|}
\end{align}
Here, $|\cX|$ denotes the cardinality of the set $\cX$, and the term subtracted from 1 can be identified as the Jaccard coefficient between $\cU(w)$ and $\cV(w)$.
We select the top scoring $w$ according to \eqref{eq:freq} and re-rank them by \eqref{eq:diversity} to select top-$k$ pivots.
Finally, we generate a set of tuples, $\cS_{\rm div}(w) = \{(w,u,v) | u \in \cU(w), v \in \cV(w)\}$, by pairing each selected pivot $w$ with its anchors in $C_1$ and $C_2$ for the purpose of filling the templates to generate prompts.

\subsubsection{Context-based Tuple Selection}
\label{sec:context}

The anchor words used in both frequency- and diversity-based tuple selection methods use PMI to measure the association between a pivot and an anchor.
This approach has two important drawbacks.

First, the number of sentences in a snapshot of a corpus taken at a specific time point could be small.
Therefore, the co-occurrences (measured at sentence level) between a pivot and a candidate anchor could be small, leading to data sparseness issues.
PMI is known to overestimate the association between rare words.\footnote{For example, if $p(w,x) \approx p(w)$. Then, \eqref{eq:pmi} reduces to $-\log p(x)$, which becomes larger for rare $x$ (i.e. when $p(x) \rightarrow 0$).}
Second, PMI considers only the two words (i.e pivot and the anchor) and \emph{not} the other words in their contexts.

We address the above-mentioned limitations of PMI by using contextualised word embeddings, $M(x,d)$ obtained from an MLM $M$ representing a word $x$ in a context $d$.
We use sentences as the contexts of words and represent a word $x$ by an embedding $\vec{x}$, computed as the average of $M(x,d)$ over $\cD(x)$, given by \eqref{eq:x}. 
\begin{align}
    \label{eq:x}
    \vec{x} = \frac{1}{|\cD(x)|} \sum_{d \in \cD(x)} M(x,d)
\end{align}
Using \eqref{eq:x}, for each word $x$ we compute two embeddings $\vec{x}_1$ and $\vec{x}_2$ respectively in $C_1$ and $C_2$.
If the word $x$ is split into multiple subtokens, we use the average of those subtoken embeddings as $\vec{x}$.
If $x$ does not exist in a particular snapshot, it will be represented by a zero vector in that snapshot.

Specifically, given $w \in C_1 \cap C_2$, $u \in C_1$ and $v \in C_2$, we score a tuple $(w,u,v)$ as in \eqref{eq:cont}.
\begin{align}
    \label{eq:cont}
    g(\vec{w}_1, \vec{u}_1) + g(\vec{w}_2, \vec{v}_2) - g(\vec{w}_2, \vec{u}_2) - g(\vec{w}_1, \vec{v}_1)
\end{align}
Here, $g(\vec{x},\vec{y})$ is the cosine similarity between the embeddings $\vec{x}$ and $\vec{y}$. 
Note that \eqref{eq:cont} assigns higher scores to tuples $(w,u,v)$ where $w$ and $u$ are highly related in $C_1$ and $w$ and $v$ in $C_2$, whereas it discourages the associations of $w$ and $u$ in $C_2$ and $w$ and $v$ in $C_1$.
This enforces the diversity requirement discussed in \autoref{sec:NN} and makes the tuple scoring method asymmetric between $C_1$ and $C_2$, which is desirable.
Finally, we rank tuples by the scores computed using \eqref{eq:cont} and select the set, $\cS_{\rm cont}$, of top-$k$ ranked tuples to fill the templates to generate prompts.

% explain how this method addresses the limitations mentioned above.
This embedding-based tuple selection method overcomes the limitations of PMI discussed at the beginning of this section as follows.
We can use contextualised embeddings from an MLM that is trained on a much larger corpus than two snapshots to obtain $M(x,d)$, thereby computing non-zero cosine similarities even when a pivot and an anchor \emph{never} co-occurs in any sentence in a snapshot. 
Moreover, contextualised embeddings are known to encode semantic information that is useful to determine the word senses~\cite{yi-zhou-2021-learning} and semantic variations~\cite{giulianelli-etal-2020-analysing}, thus enabling us to better estimate the suitability of tuples.

\subsection{Prompts from Automatic Templates}
\label{sec:auto}

Given two snapshots $C_1$, $C_2$ of a timestamped corpus and a set $\cS$ of tuples $(w,u,v)$ extracted from any one of the three methods described in \autoref{sec:tuple-selection}, we propose a method to automatically learn a diverse set of templates.
For this purpose, we can use any of the sets of tuples $\cS_{\rm freq}$, $\cS_{\rm div}$ or $\cS_{\rm cont}$ extracted as $\cS$.
We model template generation as an instance of text-to-text transformation.
For example, given the context ``\textbf{mask} \textcolor{red}{\emph{is associated with}} \textbf{hide} \textcolor{red}{\emph{in}} 2010 \textcolor{red}{\emph{and associated with}} \textbf{vaccine} \textcolor{red}{\emph{in}} 2020'', containing the tuple (\emph{mask}, \emph{hide},\emph{vaccine}), we would like to generate the sequences shown in red italics as a template.
Given a tuple $(w,u,v)$, we extract two sentences $S_1 \in C_1$ and $S_2 \in C_2$ containing the two anchors respectively $u$ and $v$, and use a pretrained T5~\cite{JMLR:v21:20-074} model to generate the slots $Z_1, Z_2, Z_3, Z_4$ for the conversion rule $\cT_g(u,v,T_1,T_2)$ shown in \eqref{eq:trans}.
\par\nobreak
{\small
\begin{align}
    \label{eq:trans}
    S_1,S_2 \rightarrow S_1 \ \langle Z_1 \rangle \ u \ \langle Z_2 \rangle \ T_1 \ \langle Z_3 \rangle \ v \ \langle Z_4 \rangle \ T_2 \ S_2
\end{align}
}%
The length of each slot to be generated is not required to be predefined, and we generate one token at a time until we encounter the next non-slot token (i.e. $u, T_1, v, T_2$).

The templates we generate must cover \emph{all} tuples in $\cS$.
Therefore, when decoding we prefer templates that have high log-likelihood values according to \eqref{eq:likelihood}.
\par\nobreak
{\small
\begin{align}
    \label{eq:likelihood}
    \vspace{-3mm}
    \sum_{i=1}^{|\cT|}\sum_{(w,u,v) \in \cS} \!\! \log P_{T5}(t_i|t_1, \ldots, t_{i-1}; \cT_g(u,v,T_1,T_2))
\end{align}
}
where $t_1,\ldots,t_{|\cT|}$ are the template tokens belonging to the slots $Z_1, Z_2, Z_3$ and $Z_4$.\footnote{Each slot can contain zero or more template tokens.}

Following \newcite{gao-etal-2021-making}, we use beam search with a wide beam width (e.g. 100) to obtain a large set of diverse templates. 
We then select the templates with the highest log-likelihood scores according to \eqref{eq:likelihood} as \emph{auto} templates.
By substituting the tuples in $\cS$ in auto templates, we generate a set of auto prompts. 
\subsection{Examples of Prompts}
\label{sec:prompt-examples}

\begin{table*}[t]
    \centering
    \small
    \begin{tabular}{ll}
    \toprule
    Template     & Type  \\
    \midrule
    \textit{$\langle w \rangle$ is associated with $\langle u \rangle$ in $\langle T_1 \rangle$, whereas it is associated with $\langle v \rangle$ in $\langle T_2 \rangle$.}     & Manual\\
   \textit{Unlike in $\langle T1 \rangle$, where $\langle u \rangle$ was associated with $\langle w \rangle$, in $\langle T2 \rangle$ $\langle v \rangle$ is associated with $\langle w \rangle$.}& Manual\\
   \textit{The meaning of $\langle w \rangle$ changed from $\langle T_1 \rangle$ to $\langle T_2 \rangle$ respectively from $\langle u \rangle$ to $\langle v \rangle$.}& Manual\\
  \textit{$\langle u \rangle$ in $\langle T_1 \rangle$ $\langle v \rangle$ in $\langle T_2 \rangle$}& Automatic\\
  \textit{$\langle u \rangle$ in $\langle T_1 \rangle$ and $\langle v \rangle$ in $\langle T_2 \rangle$}& Automatic\\
  \textit{The $\langle u \rangle$ in $\langle T_1 \rangle$ and $\langle v \rangle$ in $\langle T_2 \rangle$}& Automatic\\
  \bottomrule
    \end{tabular}
    \caption{Experimented templates. ``Manual'' denotes that the template is manually-written, whereas ``Automatic'' denotes that the template is automatically-generated.}
    \label{tbl:templates}
\end{table*}

\autoref{tbl:templates} shows the manually-written templates and the automatically learnt templates %(using the method described in \S~3.3 in the main paper). 
We see that prompts describing diverse linguistic patterns expressing how a word's usage could have changed from one time stamp to another are learnt by the proposed method.
Moreover, from \autoref{tbl:templates}, we see that automatically learnt templates tend to be shorter than the manually-written templates.
Recall that the automatic template generation method prefers sequences with high likelihoods.
On the other hand, longer sequences tend to be rare and have low likelihoods.
Moreover, we require automatic templates to cover many tuples that are selected by a particular tuple selection method, thus producing more generalisable prompts.
We believe the preference to generate shorter templates by the proposed method is due to those reasons.
\subsection{Time-adaptation by Fine-Tuning}
\label{sec:FT}

Given a set of prompts obtained by using the tuples in $\cS_{\rm freq}, \cS_{\rm div}$, or $\cS_{\rm cont}$ to fill the slots in either manually-written or automatically generated templates, we fine-tune a pretrained MLM $M$ on those prompts such that $M$ captures the semantic variation of a word $w$ from $T_1$ to $T_2$.
For this purpose, we add a language modelling head on top of $M$, randomly mask out one token at a time from each prompt, and require that $M$ correctly predicts those masked out tokens from the remainder of the tokens in the context.
We also experimented with a variant where we masked out only the anchor words from a prompt, but did not observe a notable difference in performance over random masking of all tokens.

\section{Experiments and Results}
\label{sec:exp}

\paragraph{Datasets:} We use the following four datasets that were collected and used by \newcite{hofmann-etal-2021-dynamic} for evaluating DCWEs:
\textbf{Yelp}, \textbf{Reddit}, \textbf{ArXiv}, and \textbf{Ciao}~~\cite{tang2012mtrust}.
Details of these datasets and all pre-processing steps are detailed in \autoref{sec:datasets}.
We remove duplicates as well as texts with less than 10 words in each dataset. 
We then randomly split each snapshot of a dataset into training, development, and test sets, containing respectively 70\%, 10\% and 20\% of the original dataset.

\paragraph{Evaluation Metric:}
If an MLM is correctly adapted to a timestamp $T_2$, it should be able to assign higher probabilities to the masked out tokens in unseen texts in $C_2$, sampled at $T_2$.
We follow prior work on DCWE~\cite{hofmann-etal-2021-dynamic} and  use the masked language modelling perplexity as our evaluation metric on test texts in $C_2$.
If an MLM is well-adapted to $T_2$, it will have a lower perplexity for test texts in $C_2$.

\paragraph{Baselines:}
To put our results into perspective, we consider the following baselines:
\\\noindent\textbf{Original BERT:} 
We use pretrained \texttt{BERT-base-uncased}\footnote{\url{https://huggingface.co/bert-base-uncased}} as the MLM without any fine-tuning to be consistent with \cite{hofmann-etal-2021-dynamic}. 
Further evaluations on RoBERTa~\cite{RoBERTa} are given in \autoref{sec:roberta}.
\\\noindent\textbf{BERT($T_1$):}
We fine-tune the Original BERT model on the training data sampled at $T_1$.
\\\noindent\textbf{BERT($T_2$):}
We fine-tune the Original BERT model on the training data sampled at $T_2$.
Note that this is the same training data that was used for selecting tuples in \autoref{sec:tuple-selection}
\\\noindent\textbf{FT:} The BERT models fine-tuned by the proposed method. 
%We consider the best among the different tuple selection methods. %DB: EACL-23 reviewer raised concerns about selecting the best model performance, so I have commented this out.
We use the notation FT(\texttt{model}, \texttt{template}) to denote the model obtained by fine-tuning  a given MLM using a \texttt{template}, which is either manually-written (\emph{manual}) or automatically-generated (\emph{auto}) as described in \autoref{sec:auto}.

\paragraph{Hyperparameters:}
We use the held-out development data to tune all hyperparameters.
We follow the recommendations of~\newcite{mosbach2021on} for fine-tuning BERT on small datasets and use weight decay ($0.01$) with bias correction in Adam optimiser~~\cite{Kingma:ICLR:2015}.
We use a batch size of $4$, learning rate of $3 \times 10^{-8}$, the number of tuples used for prompt-based fine-tuning ($k$) is selected from $\in \{ 500, 1000, 2000, 5000, 10000\}$, and the number of epochs is set to $20$.
(further details on hyperparameters are given in \autoref{sec:hyperparameters}).

We used a single NVIDIA RTX A6000 and 64 GB RAM in our experiments.
It takes approximately 6 hours to fine-tune, validate and test all methods reported in the paper for the four datasets. 
The run time varies depending on the number of tuples used in the proposed method. Tuple selection takes, on average, 0.5 hours. 

\subsection{Results}
\label{sec:results}

\begin{table}[t]
\centering
%\small
\resizebox{0.48\textwidth}{!}{
\begin{tabular}{lrrrr}
\toprule
MLM     & Yelp & Reddit & ArXiv & Ciao\\ 
\midrule
Original BERT          & 15.125             & 25.277     &11.142    &12.669\\
FT(BERT, Manual)       & 14.562             & 24.109    &\textbf{10.849}     &\textbf{12.371}\\
FT(BERT, Auto)        & \textbf{14.458 }    & \textbf{23.382}   &10.903     &12.394\\
\midrule
BERT($T_1$)             & 5.543           & \textbf{9.287}  &5.854    &7.423       \\
FT(BERT($T_1$), Manual) & \textbf{5.534}    &  9.327   &\textbf{5.817}    &\textbf{7.334}         \\
FT(BERT($T_1$), Auto)   & 5.541         & 9.303     &5.818      &7.347     \\
\midrule
BERT($T_2$)             & 4.718           & 8.927     &3.500      &5.840   \\
FT(BERT($T_2$), Manual) & 4.714         & \textbf{8.906}$^\dagger$  &3.499    &\textbf{5.813}$^\dagger$        \\
FT(BERT($T_2$), Auto)   & \textbf{4.708}$^\dagger$        & 8.917  &\textbf{3.499}$^\dagger$    &5.827        \\
\bottomrule
\end{tabular}
}
\caption{Masked language modelling perplexities (lower the better) on test sentences in $C_2$ in YELP, Reddit, ArXiv, and Ciao datasets are shown for different MLMs.
Best results in each block (methods using the same baseline MLM) are shown in bold, while overall best results are indicated by $\dagger$.}
\label{tbl:baselines}
\end{table}

In \autoref{tbl:baselines}, we compare the effect of fine-tuning BERT MLMs using the prompts generated by filling the selected tuples in either the manually-written (manual) or automatically learnt (auto) templates.
We use the optimal tuples selection method and the number of tuples for prompt-based fine-tuning, decided using the validation data in each datasets.

From \autoref{tbl:baselines} we see that the \textbf{Original BERT} model has the highest perplexity scores in all four datasets.
This shows that the \textbf{Original BERT} model is not accurately capturing the meanings of words as used in $C_2$.
Although fine-tuning \textbf{Original BERT} using manual or auto prompts improves the perplexity scores, they still remain very high.
\textbf{BERT($T_1$)}, obtained by fine-tuning the \textbf{Original BERT} model on $C_1$, immediately reduces the perplexity scores on both datasets.
Recall that $C_1$ is sampled from the same domain but at $T_1$, which is different from $T_2$. 
This indicates the importance of using in-domain data for fine-tuning MLMs eventhough it might be from a different timestamp. 

On the other hand, \textbf{BERT($T_2$)}, obtained by fine-tuning \textbf{Original BERT} on the training data from $T_2$, further reduces perplexity over \textbf{BERT($T_2$)}.
Importantly, fine-tuning using both manual and auto prompts further reduces perplexity over \textbf{BERT($T_2$)}.
Overall, the best performances on Yelp and ArXiv are reported by fine-tuning \textbf{BERT($T_2$)} using auto prompts (i.e. \textbf{FT(BERT($T_2$), Auto)}), whereas the same on Reddit and Ciao are reported by manual prompts (i.e. \textbf{FT(BERT($T_2$), Manual)}).
Applying auto or manual prompts for fine-tuning \textbf{BERT($T_1$)} improves perplexity in Yelp, ArXiv, and Ciao, but not on Reddit. 
This shows the importance of first fine-tuning BERT on $C_2$ (in-domain and contemporary) before using prompts to further fine-tune models, because prompts are designed or learnt for the purpose for adapting to $T_2$ and not to $T_1$, reflecting the direction of the time adaptation ($T_1 \rightarrow T_2$).

\begin{table}[t]
\centering

\resizebox{0.48\textwidth}{!}{
\begin{tabular}{lrrrr}
\toprule
MLM    &    Yelp & Reddit & ArXiv & Ciao\\
\midrule
FT(BERT($T_2$), Manual) & 4.714         & \textbf{8.906}$^\dagger$  &3.499    &\textbf{5.813}$^\dagger$ \\
FT(BERT($T_2$), Auto)   & \textbf{4.708}$^\dagger$        & 8.917  &\textbf{3.499}$^\dagger$    &5.827 \\
\midrule
TempoBERT & 5.516   & 12.561 & 3.709    & 6.126 \\
CWE             &4.723      &\textbf{9.555}      &3.530      &5.910\\
DCWE [temp. only]  &4.723     &9.631      &3.515      &\textbf{5.899}\\
DCWE [temp.+social] &\textbf{4.720}     &9.596      &\textbf{3.513}      &5.902 \\
\bottomrule
\end{tabular}
}
\caption{MLM perplexities (lower the better) are shown for the proposed method, previously proposed TempoBert~\cite{TempoBERT}, and DCWE variants~\cite{hofmann-etal-2021-dynamic}. Best results in each block (methods using the same baseline MLM) are shown in bold, while overall best results are indicated by $\dagger$.}
\label{tbl:sota}
\end{table}

Although there is prior work on non-contextualised dynamic embeddings, we cannot perform language modelling with those as the probability of predicting a word will be a constant independent of its context.
Moreover, none of those work evaluate on the benchmarks proposed by \newcite{hofmann-etal-2021-dynamic}, which we also use.
Therefore, we consider the current SoTA for DCWEs proposed by \newcite{hofmann-etal-2021-dynamic} and the SoTA for time-adaptation, TempoBERT~\cite{TempoBERT}, as our main comparison points in \autoref{tbl:sota}.

The DCWE method proposed by \newcite{hofmann-etal-2021-dynamic} uses BERT fine-tuned on training data from $T_2$ as the baseline MLM (i.e. \textbf{CWE}).
Moreover, their method is available in two flavours: a version that uses both social and temporal information (denoted as \textbf{DCWE [social + temp.}]) and a version where social information is ablated (denoted as \textbf{DCWE [temp.]}).
Given that we do not use social information in our proposed method, the direct comparison point for us would be \textbf{DCWE [temp.]}.
We see that our proposed method with both manual and auto prompts consistently outperforms both flavours of the SoTA DCWEs proposed by \newcite{hofmann-etal-2021-dynamic} in all datasets.

TempoBert inserts a special token indicating the time period at the beginning of each sentence in a training corpus, and fine-tunes BERT on the corpora available for each time period.
We trained TempoBert on our datasets for the same number of epochs as and with an initial learning rate of 3e-6 and measured perplexity on the same test splits.
As seen from \autoref{tbl:sota}, the proposed method using both manual and automatic templates outperforms TempoBert in all datasets.

\begin{figure*}[t]
\centering
\subfigure{
\includegraphics[width=7.5cm,valign=c]{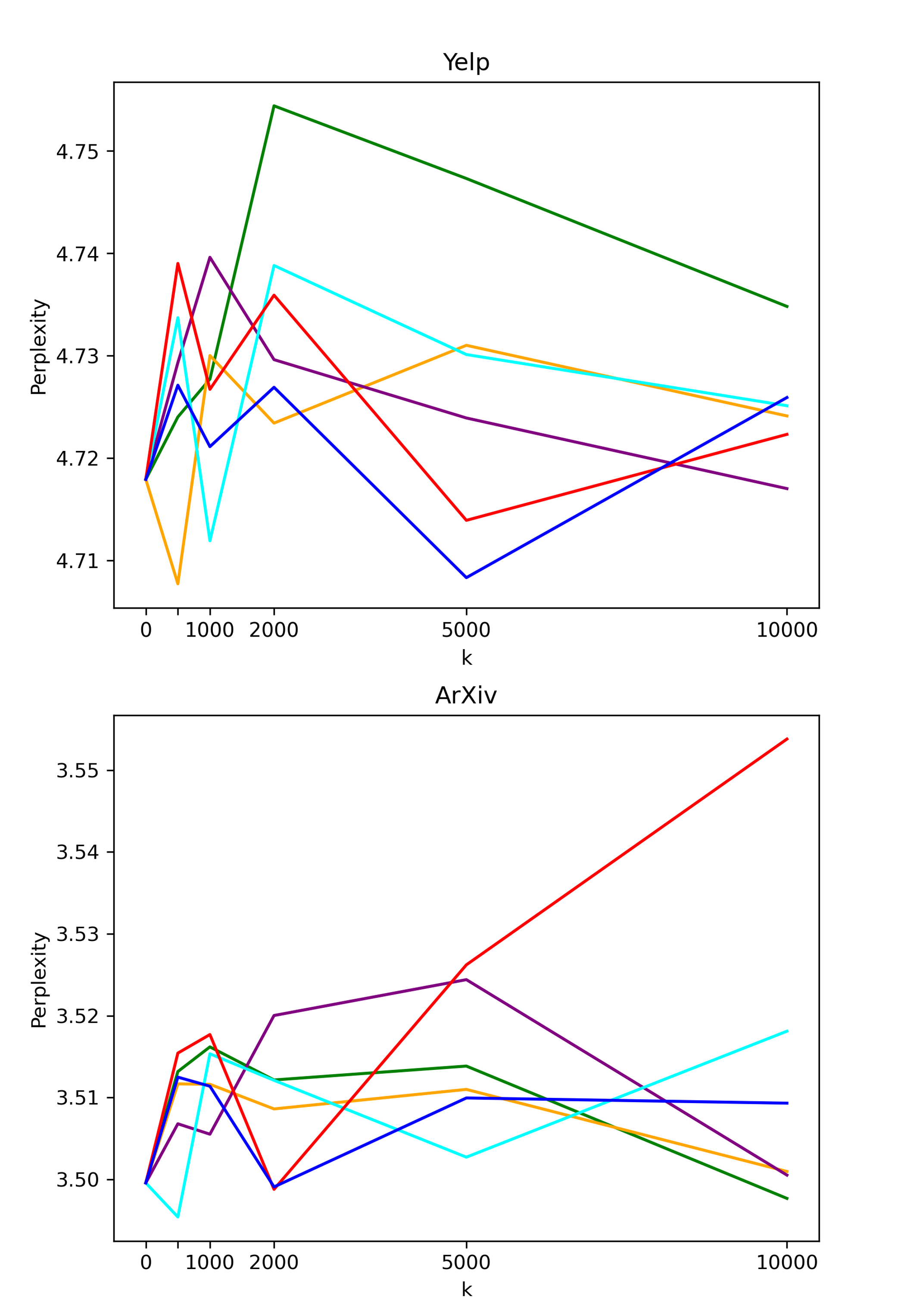}
%\caption{fig1}
}
\quad
\subfigure{
\includegraphics[width=7.5cm,valign=c]{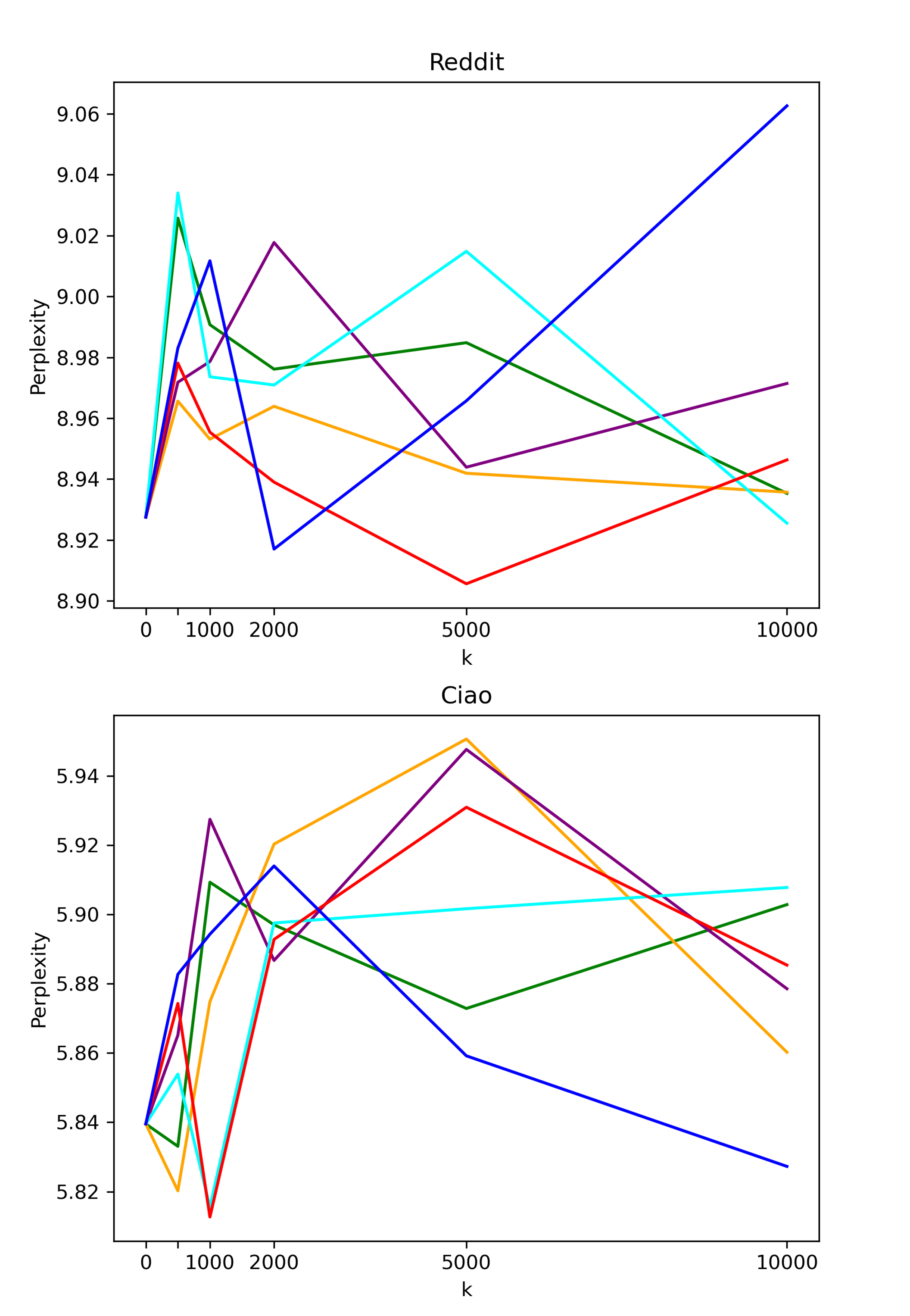}
}
\quad
\subfigure{
\includegraphics[width=14cm,valign=c]{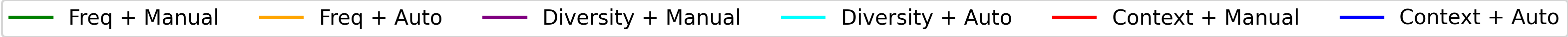}
}
\caption{Adapting BERT($T_2$) to $T_2$ on Yelp (top left), Reddit (top right), ArXiv (bottom left), and Ciao (bottom right) datasets using different tuple selection methods (\textbf{Freq}uency, \textbf{Diversity}, \textbf{Context}) and templates (\textbf{Auto}, \textbf{Manual}).  Perplexity scores are shown against the the number of tuples ($k$) used in prompt-based fine-tuning.}
\label{fig:T2}
\vspace{-3mm}
\end{figure*}

The number of tuples selected (i.e. $k$) to generate prompts with manually-written or automatically generated templates determines the number of prompts used to fine-tune a given MLM.
To study the effect of $k$, we use a particular tuple selection method and select the top-ranked $k$ tuples according to that method with either a manually-written (\textbf{Manual}) or automatically learnt (\textbf{Auto}) template to generate prompts.
This results in six different types of prompts.
Next, we fine-tune a given MLM using the generated prompts and repeat this process for increasing $k$ values.
\autoref{fig:T2} shows the results of fine-tuning \textbf{BERT($T_2$)} to $T_2$. 
(Results for \textbf{BERT($T_1$)} are shown in \autoref{sec:tune-C1})

Overall, from \autoref{fig:T2} we see that when the number of tuples used for fine-tuning increases, almost all methods reach some minimum perplexity score.
However, we see that for each method, its minimum perplexity scores on different datasets is obtained by using different $k$ tuples.
Recall that each tuple selection method ranks tuples by some goodness score. 
Therefore, when we increase $k$ we are using less reliable noisy tuples to generate prompts, thus leading to reduced performance. 
Interestingly, we see that the best performances can be obtained with relatively a smaller number of tuples ($<5000$) in all datasets.

A closer look reveals that on Yelp, all \textbf{Freq+Auto} and \textbf{Context+Auto} obtain similar best performances.
However, \textbf{Freq+Auto} reaches its optimal point with 500 tuples, whereas \textbf{Context+Auto} requires 5000 tuples.
Among the three tuple selection methods Context-based tuple selection is the best.
Frequency-based tuple selection method works well with auto templates but not so much with manual ones. 
This shows that auto templates can be used to extract good performance from a simple tuple selection method such as Frequency-based tuple selection.

On Reddit, the overall best performance is obtained by \textbf{Context+Manual} with 5000 tuples, and its performance drops with higher $k$ values, due to the increasing noise in tuples as explained above.
Likewise in Yelp, context-based tuple selection emerges as the overall best method in Reddit as well with 5000 tuples.
However, context-based tuple selection has to be used with manual templates to obtain good performance on Reddit, whereas in Yelp using it with the auto templates was the best.

On ArXiv, both \textbf{Freq+Manual} and \textbf{Diversity+Auto} reach similar best performances. While \textbf{Freq+Manual} requires 500 tuples, it takes \textbf{Diversity+Auto} 10000 tuples to reach its best performance.
Unlike Yelp and Reddit, the best tuple selection in ArXiv is Diversity-based tuple selection. The Frequency-based tuple selection also similar performance, but requires more tuples. For Context-based tuple selection, although it improves the perplexity scores over the baseline MLM, the improvements are smaller than other methods.

On Ciao, \textbf{Context+Manual} and \textbf{Diversity + Auto} obtain similar best performances, both with 1000 tuples.
Similarly as Yelp and Reddit, the overall best tuple selection is Context-based tuple selection, which obtains the best perplexity scores. Diversity-based tuple selection also has good performance, although it only occurs when it is used with auto templates.  

\begin{table}[t]
    \centering
    \small
    \begin{tabular}{p{2cm} p{4cm}}
    \toprule
    Pivot ($w$) & Anchors (\red{$u$}, \blue{$v$}) \\
    \midrule
    \emph{place} &  (\red{burgerville}, \blue{takeaway}), (\red{burgerville}, \blue{dominos}), (\red{joes}, \blue{dominos})\\
    \emph{service} & (\red{doorman}, \blue{staffs}), (\red{clerks}, \blue{personnel}), (\red{clerks}, \blue{administration}) \\
    \emph{phone} & (\red{nokia}, \blue{iphone}), (\red{nokia}, \blue{ipod}), (\red{nokia}, \blue{blackberry}) \\
    \bottomrule
    \end{tabular}
    \caption{Top-ranked pivots $w$ and their associated anchors $u$ and $v$ selected according to the contextualised tuple selection method from Yelp (Row 1 and 2) and Ciao (Row 3).}
    \label{tbl:tuples}
    \vspace{-5mm}
\end{table}

\autoref{tbl:tuples} shows some examples of the top scoring pivots and their anchors retrieved by the context-based tuple selection method from Yelp and Ciao.
From Yelp, in year 2010 ($T_1$), we see that dine-in restaurants such as \emph{burgerville}\footnote{\url{www.burgerville.com}} and \emph{joes}\footnote{\url{www.joespizzanyc.com}} are associated with \emph{place}, whereas in 2020 \emph{takeaway} and \emph{dominos}\footnote{\url{www.dominos.com}} are associated with \emph{place} probably due to the COVID-19 imposed lockdowns restricting eating out.
Moreover, we observe a shift in office-related job titles between these time periods where \emph{service} is closely associated with \emph{doorman} (T1: 108, T2: 48) and \emph{clerks} (T1: 115, T2: 105), which are rarely used in 2020 and are replaced with more gender-neutral titles such as \emph{staff} (T1: 28618, T2: 60421), \emph{personnel} (T1: 85, T2: 319) and \emph{administration} (T1: 37, T2: 109). From Ciao, in year 2001 ($T_1$), we see that phone brands like \emph{nokia}\footnote{\url{www.nokia.com}} are closely connected with \emph{phone}, while in year 2011 ($T_2$), as companies such as apple\footnote{\url{www.apple.com}} and \emph{blackberry}\footnote{\url{www.blackberry.com}} took a large part of the mobile phone market, \emph{iphone}, \emph{ipod}, and \emph{blackberry} become more related with \emph{phone}.

\section{Conclusion}
We propose an unsupervised method to learn DCWEs by time-adapting a pretrained MLM using prompts from manual and automatic templates.
Experimental results on multiple datasets demonstrate that our proposed method can obtain better perplexity scores on unseen target texts compared to prior work.
In the future, we plan to extend the proposed method to adapt multilingual word embeddings.

\section{Limitations}
This paper takes into account the temporal semantic variations of words and proposes a method to learn dynamic contextualised word embeddings by time-adapting an MLM using prompt-based fine-tuning methods. 
In this section, we highlight some of the important limitations of this work.
We hope this will be useful when extending our work in the future by addressing these limitations.

The learned dynamic contextualised word embeddings are limited to the English language, which is a morphologically limited language. 
Therefore, the findings reported in this work might not generalise to other languages.
However, there are already numerous multilingual MLMs such as mBERT~\cite{devlin-etal-2019-bert}, XLM~\cite{CONNEAU2019} and XLM-R~\cite{conneau2020unsupervised}, to name a few. 
Extending our work to multilingual dynamic contextualised word embeddings will be a natural line of future work.

Dynamic contextualised word embeddings represent words as a function of extralinguistic context~\cite{hofmann-etal-2021-dynamic}, which consider both time and social aspects of words.
However, in this paper we focused solely on the temporal aspect and ignored the social aspect. 
Extending our work to take into account the social semantic variations of a word is an important future direction. 

Due to the costs involved when fine-tuning large-scale MLMs, we keep the number of manually-written and automatically learnt templates to a manageable small number as shown in \autoref{tbl:templates} in \autoref{sec:prompt-examples}. 
However, it remains to be evaluated the impact of increasing the number of templates on the performance of the proposed method.

\section{Ethical Considerations}
In this paper, we considered the problem of capturing temporal semantic variation of words by learning dynamic contextualised word embeddings.
For this purpose, we proposed a method to adapt a masked language model from to a given time stamp.
We did not collect, annotate or release new datasets during this process.
However, we used pretrained MLMs and four datasets from the internet (Yelp, Reddit, Arxiv, and Ciao).
It is known that pretrained MLMs contain unfair social biases~\cite{may-etal-2019-measuring,nadeem-etal-2021-stereoset,Kaneko:EACL:2021b,Kaneko:MLM:2022}.
Such biases can be amplified by fine-tuning methods, especially when the fine-tuning prompts are extracted from social data such as customer reviews in Yelp or discussions in Reddit.
Therefore, we consider it is important to further evaluate~\cite{crows-pairs,nadeem-etal-2021-stereoset,Kaneko:AAAI2022} the adapted MLMs for social biases, and if they do exist, then apply appropriate debiasing methods~\cite{Kaneko:EACL:2021b,lauscher-etal-2021-sustainable-modular} before the MLMs are deployed in downstream NLP applications that are used by billions of users world-wide.

\section*{Acknowledgements}
Danushka Bollegala holds concurrent appointments as a Professor at University of Liverpool and as an Amazon Scholar. This paper describes work performed at the University of Liverpool and is not associated with Amazon.

\bibliography{dynamic-embd}
\bibliographystyle{acl_natbib}

\appendix
\section*{Appendix}

\section{Fine-tuning results on $C_1$}
\label{sec:tune-C1}

\begin{figure*}[h!]
\centering
\subfigure{
\includegraphics[width=7.5cm, valign=c]{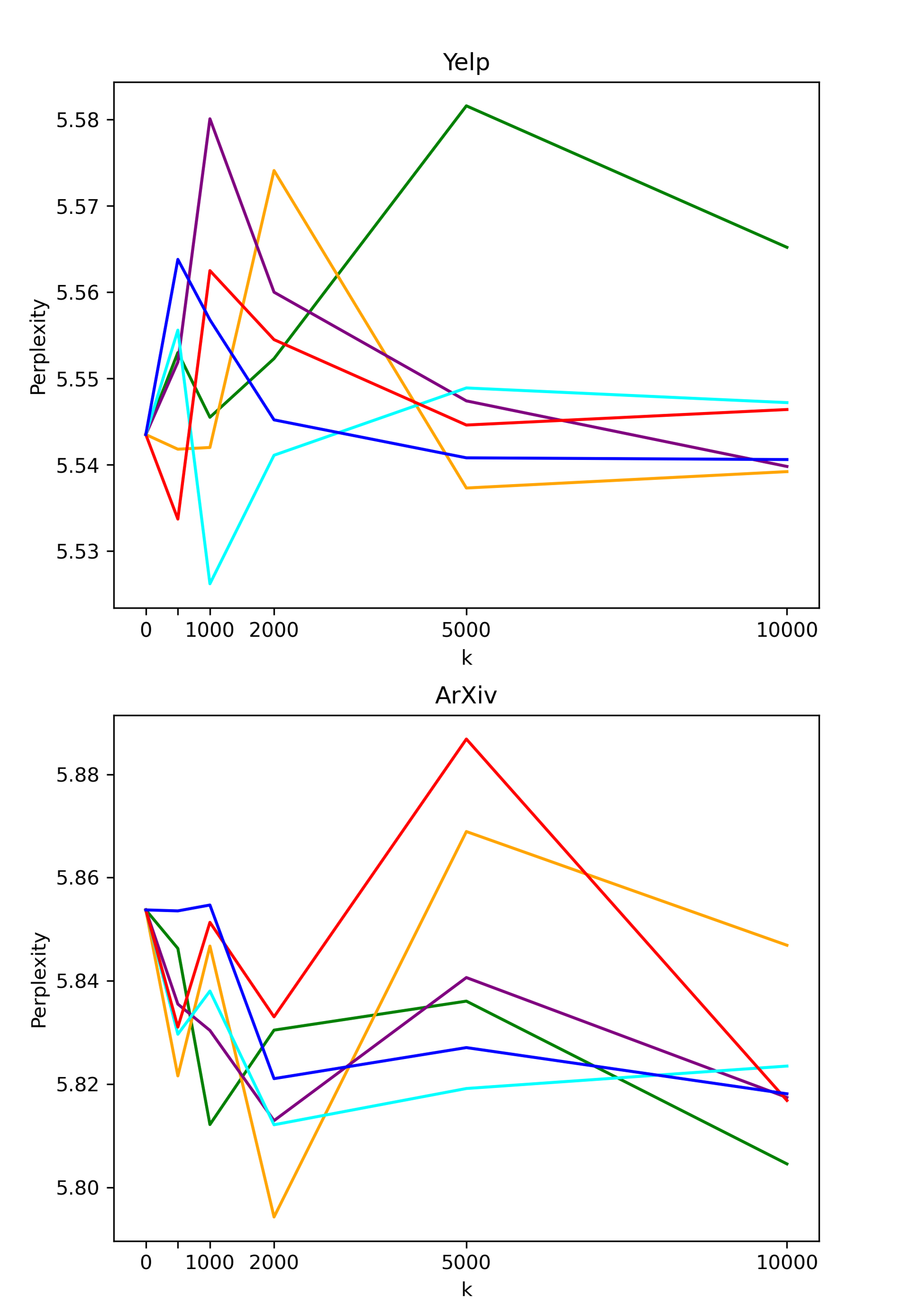}
%\caption{fig1}
}
\quad
\subfigure{
\includegraphics[width=7.5cm, valign=c]{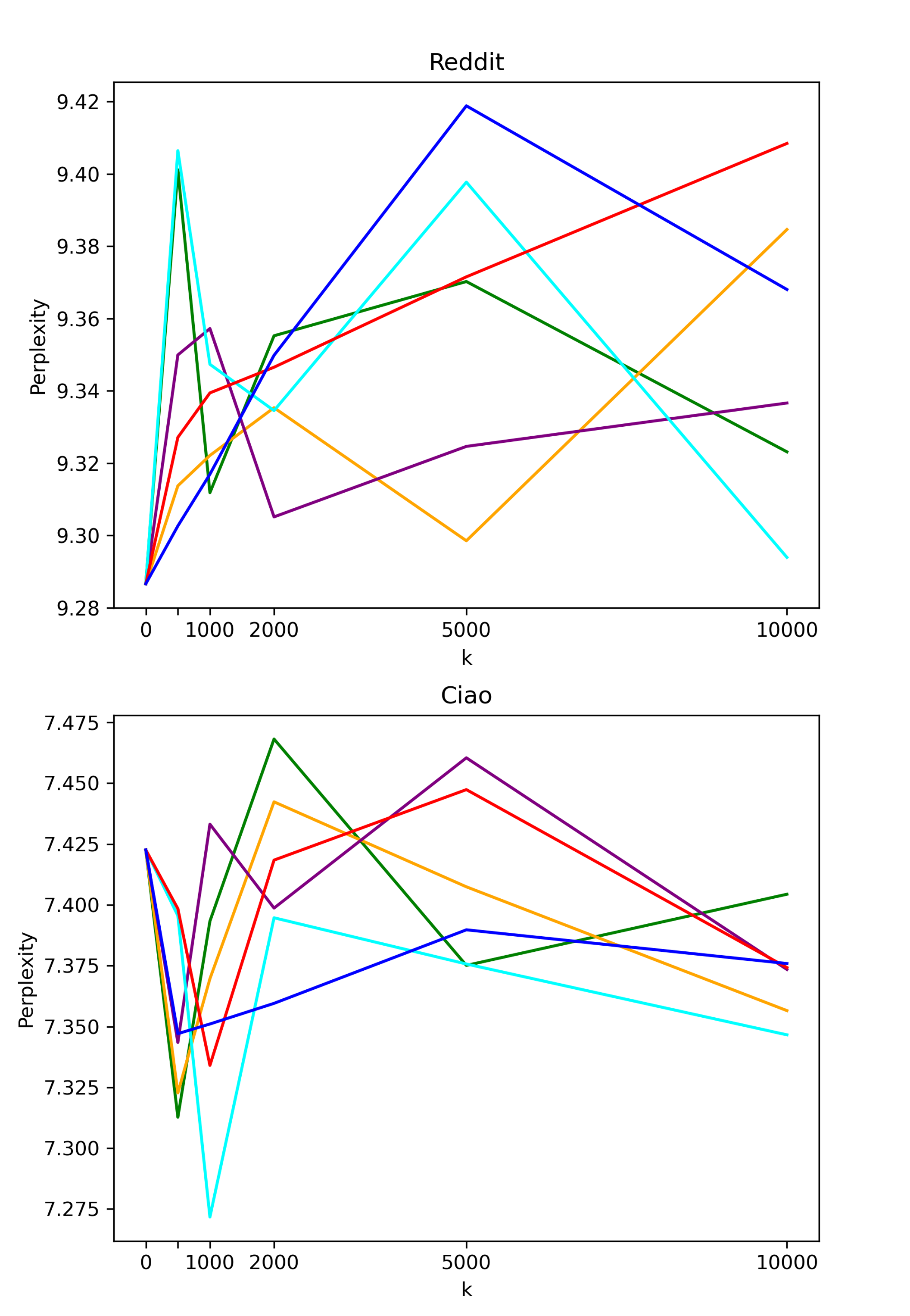}
}
\quad
\subfigure{
\includegraphics[width=14cm, valign=c]{legend.png}
}
\caption{Adapting BERT($T_1$) to $T_2$ on YELP (top left), Reddit (top right), ArXiv (bottom left) and Ciao (bottom right) datasets using different tuple selection methods (\textbf{Freq}uency, \textbf{Diversity}, \textbf{Context}) and templates (\textbf{Auto}, \textbf{Manual}).  Perplexity scores are shown against the the number of tuples ($k$) used in prompt-based fine-tuning.}
\label{fig:T1}
\end{figure*}

\autoref{fig:T1} shows the effect of the number of tuples (i.e.~$k$) selected using different tuple selection methods, and the perplexity scores for the \textbf{BERT($T_1$)} models, fine-tuned using the prompts generated by filling the slots with those tuples in either manually-written (\textbf{Manual}) or automatically-generated (\textbf{Auto}) templates.
Note that we have three methods to select tuples (i.e. \textbf{Freq}uency-based tuple selection, \textbf{Diversity}-based tuple selection, and \textbf{Context}-based tuple selection).
Combined with the two methods for obtaining tuples, we have six comparisons in \autoref{fig:T1} on Yelp, Reddit, ArXiv, and Ciao datasets.
Because a only a single template is used in each setting, the number of tuples ($k$) is equal to the number of prompts used to fine-tune an MLM in this experiment.

On Yelp, we see that \textbf{Freq+Auto} and \textbf{Diversity+Auto} both obtain the lowest (best) perplexity scores.
In particular, we see that \textbf{Freq+Auto} reaches this optimal performance point with as less as 500 prompts, whereas \textbf{Diversity+Auto} requires 1000 prompts. 
However, when we increase the number of prompts beyond the optimal performance points for each method, we see that the perplexity increases due to the added noise when using low-scoring tuples for generating prompts.
Although for both of those methods the perplexity scores drop again when a large number of prompts are being used (i.e. more than 5000 prompts) only \textbf{Diversity+Auto} recovers to the best performance level it obtained with 1000 prompts.
Therefore, we note that there is a trade-off here between the quality vs. quantity of using noisy prompts for fine-tuning.
However, from a computational point of view it is desirable to use a small number of prompts if that suffice to obtain good performance.
Therefore, we recommend using \textbf{Freq+Auto} in this case because it obtained good performance with only 500 prompts.

On Reddit we see that the perplexity increases with the number of prompts in almost all methods from the start.
However, they reach a peak and then start decreasing again.
However, among all methods we see that only Diversity+Auto recovers to its initial levels.
In fact, with 10000 prompts it is able to report perplexity scores lower than that of its initial values, thus reporting the best performance on Reddit by any fine-tuning method.
However, recall that auto templates were specifically learnt to obtain good performance when adapting from $T_1$ to $T_2$, and the perplexity scores obtained by fine-tuning \textbf{BERT($T_2$)} are much better than those obtained by fine-tuning \textbf{BERT($T_1$)} (which are shown in \autoref{fig:T1}) as explained in the main body of the paper.

On ArXiv we see that \textbf{Freq+Auto} obtain the best perplexity score.
In almost all methods, the perplexity scores drop first and then increase. However, the increases are followed by drops and then increases. The trend of perplexity scores regarding the tuple numbers seems a wave.
Unlike other mehtods, \textbf{Context+Auto} almost continues to improve its performances as the number of tuples increases.
\textbf{Freq+Auto} is the overall best method as it reaches the best perplexity score with 2000 tuples. In addition, we see that the potential performances of \textbf{Context+Auto} would be high since its performances increase with the number of tuples.

On Ciao we see that \textbf{Diversity+Auto} obtains the best perplexity score and it is much better than other methods.
Unlike other datasets, all methods reach their best perplexity scores with small numbers of tuples ($<1000$).
The trend of perplexity score changing regarding the numbers of tuples is almost the same in all methods: drop, increase, and drop.

\section{Experiment on RoBERTa}
\label{sec:roberta}

\begin{figure*}[t]
\centering
\subfigure{
\includegraphics[width=7.5cm, valign=c]{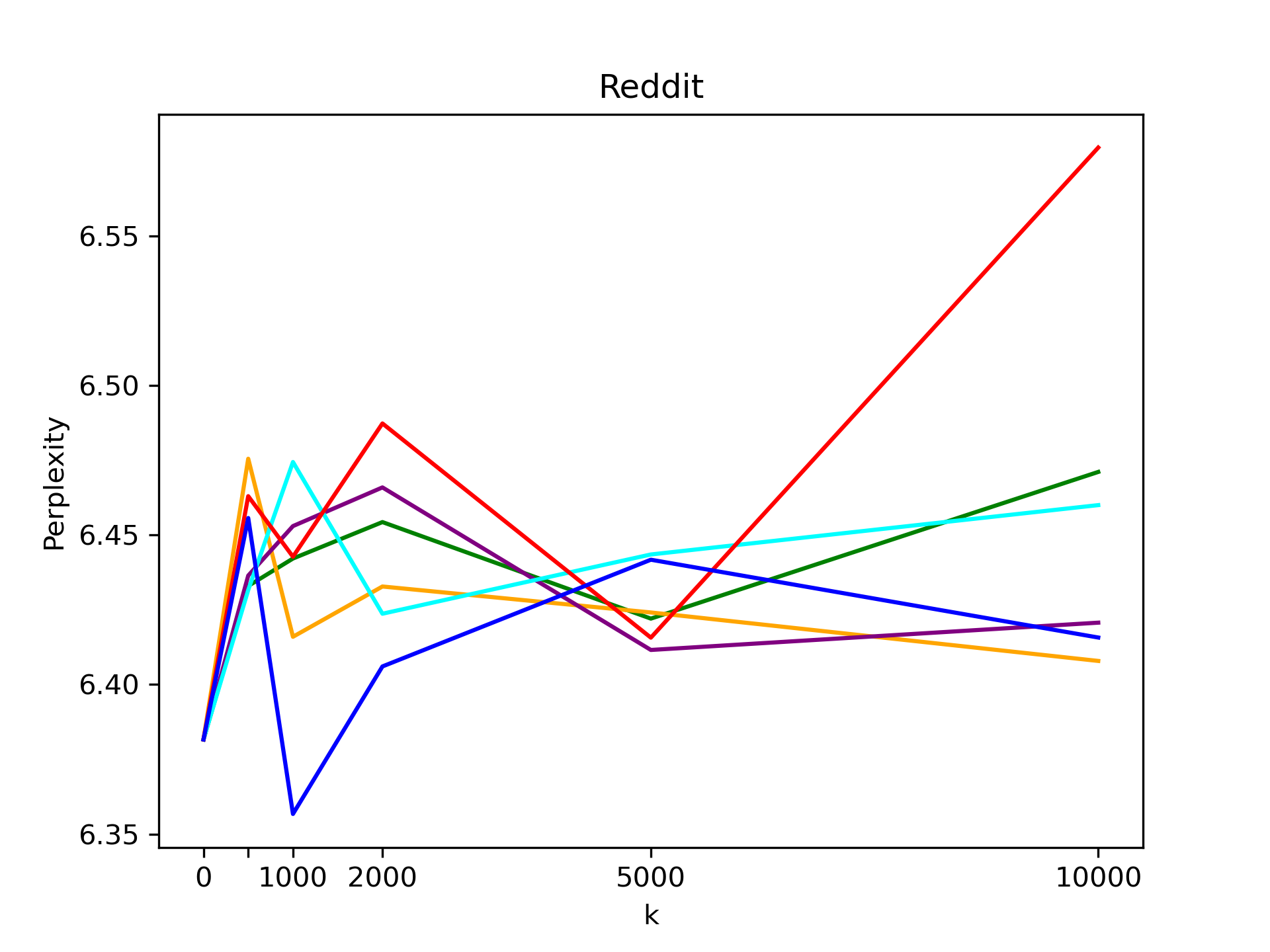}
%\caption{fig1}
}
\quad
\subfigure{
\includegraphics[width=7.5cm, valign=c]{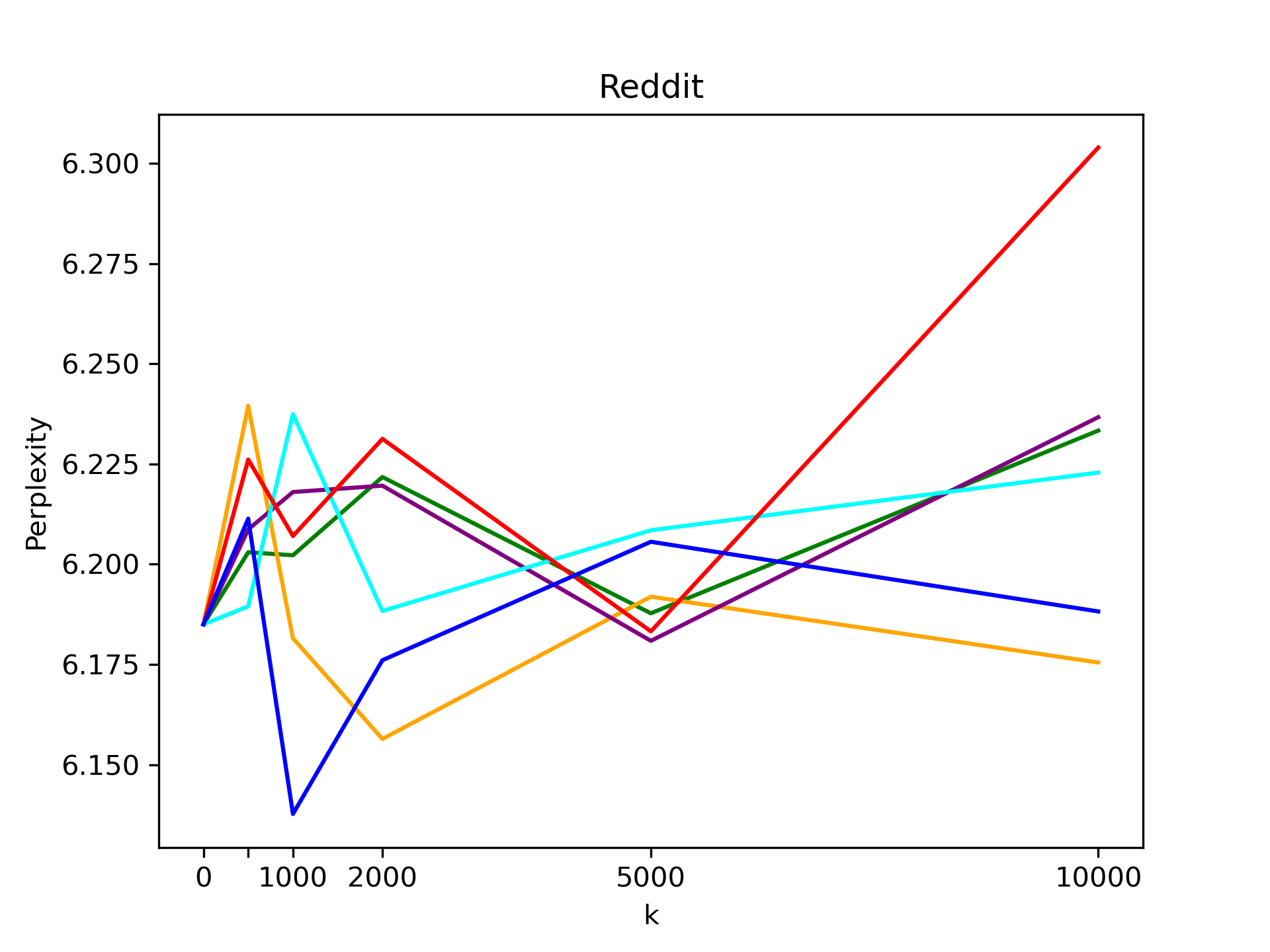}
}
\subfigure{
\includegraphics[width=14cm, valign=c]{legend.png}
}
\caption{Adapting RoBERTa($T_1$) to $T_2$ (left) and RoBERTa($T_2$) to $T_2$ (right) on Reddit dataset using different tuple selection methods (\textbf{Freq}uency, \textbf{Diversity}, \textbf{Context}) and templates (\textbf{Auto}, \textbf{Manual}).  Perplexity scores are shown against the the number of tuples ($k$) used in prompt-based fine-tuning.}
\label{fig:RoBERTa}
\end{figure*}

To explore the proposed method's potential on other MLMs than BERT, we conduct a small-scale experiment on RoBERTa. The baselines and evaluation metric setting are similar to the experiment in the main body except that the MLM is changed to \texttt{RoBERTa-base}\footnote{\url{https://huggingface.co/roberta-base}} and we only use the Reddit datasets.

In \autoref{tbl:roberta} we compare the effect of fine-tuning RoBERTa MLMs using the prompts from both automatic and manual templates. From \autoref{tbl:roberta} we see that the \textbf{Original RoBERTa} has the highest perplexity score in Reddit dataset, and fine-tuning \textbf{Original RoBERTa} with manual or auto prompts improves the perplexity. While applying manual prompts does not improve the perplexity score over \textbf{RoBERTa($T_1$)}, fine-tuning with auto prompts makes some improvements. Likewise the results of the main experiment on BERT, fine-tuning using both manual and auto prompts further reduces perplexity over \textbf{RoBERTa($T_2$)}.

\autoref{fig:RoBERTa} shows the results of fine-tuning \textbf{RoBERTa($T_1$)} and \textbf{RoBERTa($T_2$) to $T_2$}.

For \textbf{RoBERTa($T_1$)}, \textbf{Context+Auto} has the best perplexity score with 1000 tuples. However, the context-based tuple selection method only improve the perplexity score over the baseline when it is used with auto templates. Moreover, \textbf{Context+Auto} is the only method that improves the perplexity against the baseline MLM.

For \textbf{RoBERTa($T_2$)}, similar as \textbf{RoBERTa($T_1$)}, \textbf{Context+Auto} obtain the lowest (best) perplexity score with 1000 tuples. \textbf{Freq+Auto} also reaches a similar perplexity score with 2000 tuples. As tuple numbers increase, almost all methods first reach optimal points, and then their perplexity scores increase as the tuple numbers increase. \textbf{Context+Auto} is the overall best method because its best performance and the smallest tuple number.
\begin{table}[t]
\centering
%\small
\begin{tabular}{lr}
\toprule
MLM     Reddit\\ 
\midrule
Original RoBERTa          & 13.997            \\
FT(RoBERTa, Manual)       & 13.818             \\
FT(RoBERTa, Auto)        & \textbf{13.323}    \\
\midrule
RoBERTa($T_1$)             & 6.382         \\
FT(RoBERTa($T_1$), Manual) & 6.443    \\
FT(RoBERTa($T_1$), Auto)   & \textbf{6.357}     \\
\midrule
RoBERTa($T_2$)             & 6.185          \\
FT(RoBERTa($T_2$), Manual) & 6.183        \\
FT(RoBERTa($T_2$), Auto)   & \textbf{6.138}$^\dagger$\\
\bottomrule
\end{tabular}
\caption{Masked language modelling perplexities (lower the better) on test sentences in $C_2$ in Reddit datasets are shown for different MLMs.
Best results in each block (methods using the same baseline MLM) are shown in bold, while overall best results are indicated by $\dagger$.}
\label{tbl:roberta}
\end{table}

\section{Datasets}
\label{sec:datasets}

\paragraph{Yelp:} Yelp is a platform which provides crowd-sourced reviews on businesses. We select publicly available reviews\footnote{\url{https://www.yelp.com/dataset}} covering the years 2010 (=$T_1$) and 2020 (=$T_2$).
\paragraph{Reddit:} Reddit is a social media platform covering a wide range of topics arranged into communities called \emph{subreddits}. 
Following \newcite{hofmann-etal-2021-dynamic}, from the publicly released Reddit posts,\footnote{\url{https://files.pushshift.io/reddit/comments/}} we take all comments from September 2019 (=$T_1$) and April 2020 (=$T_2$), which reflect the effects of the COVID-19 pandemic. 
We remove subreddits with fewer than 60 comments and randomly sample 60 comments per subreddit.
\paragraph{ArXiv:} ArXiv is an open-access repository of scientific articles. We obtain abstracts of papers published at years 2001 (=$T_1$) and 2020 (=$T_2$) on ArXiv from a publicly available dataset\footnote{\url{https://www.kaggle.com/datasets/Cornell-University/arxiv}}. Following \newcite{hofmann-etal-2021-dynamic}, we drop those data under ArXiv's subjects (e.g., \textsc{CS.CL}) that has less than 100 publications between 2001 and 2020.
\paragraph{Ciao:} Ciao is a product review site. We select reviews from years 2000 (=$T_1$) and 2011 (=$T_2$) from a publicly released dataset~\cite{tang2012mtrust}\footnote{\url{https://www.cse.msu.edu/~tangjili/trust.html}}.

\section{Hyperparameters}
\label{sec:hyperparameters}

\begin{table*}[h!]
    \centering
    \small
    \begin{tabular}{l  c c c c c c c c c c c}
    \toprule
     & \multicolumn{2}{c}{Yelp}& & \multicolumn{2}{c}{Reddit} & & \multicolumn{2}{c}{ArXiv} & & \multicolumn{2}{c}{Ciao} \\
     \cline{2-3} \cline{5-6} \cline{8-9} \cline{11-12}
     MLM & $l$  & s & &  $l$  &  s& &  $l$  &  s& &  $l$  &  s\\
     \midrule
     FT(BERT($T_2$), Manual) & 3e-8 & $-$ & & 1e-8 & $-$ & & 5e-7 & warm up$^*$ & & 6e-8 & warm up\\
     FT(BERT($T_2$), Auto) & 3e-8 & $-$ & & 2e-7 & warm up$^*$ & & 3e-8 & $-$ & & 6e-7 & warm up\\
     FT(RoBERTa($T_2$), Manual) & $-$ & $-$ & & 3e-8 & $-$& & $-$ & $-$ & & $-$ & $-$\\
     FT(RoBERTa($T_2$), Auto) & $-$ & $-$ & & 3e-8 & $-$ & & $-$ & $-$ & & $-$ & $-$\\
     \bottomrule
    \end{tabular}
    \caption{Hyperparameters setting for adapting BERT($T_2$) and RoBERTa($T_2$) to $T_2$ on Yelp, Reddit, ArXiv, and Ciao datasets. Here, ``warm up'' denotes that the learning rate is linearly increased for the first $p$\% of the steps, where $p = 10$ for \{Reddit,Arxiv\} and $p=65$ for \{Ciao\} if applicable. $^*$ indicates that the learning rate is linearly decayed until zero after the ``warm up''.}
    \label{tbl:hyper}
\end{table*}

\autoref{tbl:hyper} shows the hyperparameter values for fine-tuning \textbf{BERT($T_2$)} and \textbf{RoBERTa($T_2$)} using prompts on $T_2$. We used \texttt{T5-base}\footnote{\url{https://huggingface.co/t5-base}} to generate automatic prompts. The batch size of generating process is 32, and the width of the beam search is set to 100.

To examine the influence of the random seeds, we firstly perform \textbf{FT(BERT($T_2$),Auto)} on ArXiv with different numbers of tuples and three random seeds. Then we calculated the mean values and the standard deviations of perplexities with different tuple numbers regarding the random seeds. As we average the mean values and the standard deviation, we see that the average standard deviation (i.e. 0.0066) is much smaller than the average mean (i.e. 3.5079), which is nearly 1/1000. Thus, we only use 123 as the random seed for the formal experiments.

\end{document}